\documentclass{article}
\usepackage{spconf,amsmath,graphics,epsfig,amsmath,nccmath,multirow,caption,subfig,booktabs,amssymb,bbding}

\title{OAS-Net: Occlusion Aware Sampling Network for Accurate Optical Flow}
\name{
	Lingtong Kong \quad Xiaohang Yang \quad Jie Yang
}
\address{Institute of Image Processing and Pattern Recognition\\
	Shanghai Jiao Tong University, China}
\begin{document}
%
\maketitle
\begin{abstract}
Optical flow estimation is an essential step for many real-world computer vision tasks. Existing deep networks have achieved satisfactory results by mostly employing a pyramidal coarse-to-fine paradigm, where a key process is to adopt warped target feature based on previous flow prediction to correlate with source feature for building 3D matching cost volume. However, the warping operation can lead to troublesome ghosting problem that results in ambiguity. Moreover, occluded areas are treated equally with non occluded regions in most existing works, which may cause performance degradation. To deal with these challenges, we propose a lightweight yet efficient optical flow network, named OAS-Net (occlusion aware sampling network) for accurate optical flow. First, a new sampling based correlation layer is employed without noisy warping operation. Second, a novel occlusion aware module is presented to make raw cost volume conscious of occluded regions. Third, a shared flow and occlusion awareness decoder is adopted for structure compactness. Experiments on Sintel and KITTI datasets demonstrate the effectiveness of proposed approaches.
\end{abstract}
\section{Introduction}
\label{sec:intro}
Optical flow estimation is a longstanding and fundamental task in computer vision. It plays a key role for many real-world applications, such as object tracking~\cite{1041198}, action recognition~\cite{NIPS2014_5353} and scene understanding~\cite{Hur_2016}.

\begin{figure}[t]
	\captionsetup[subfigure]{farskip=1pt}
	\centering
	\subfloat[Source Image]
	{
		\includegraphics[width=0.48\linewidth]{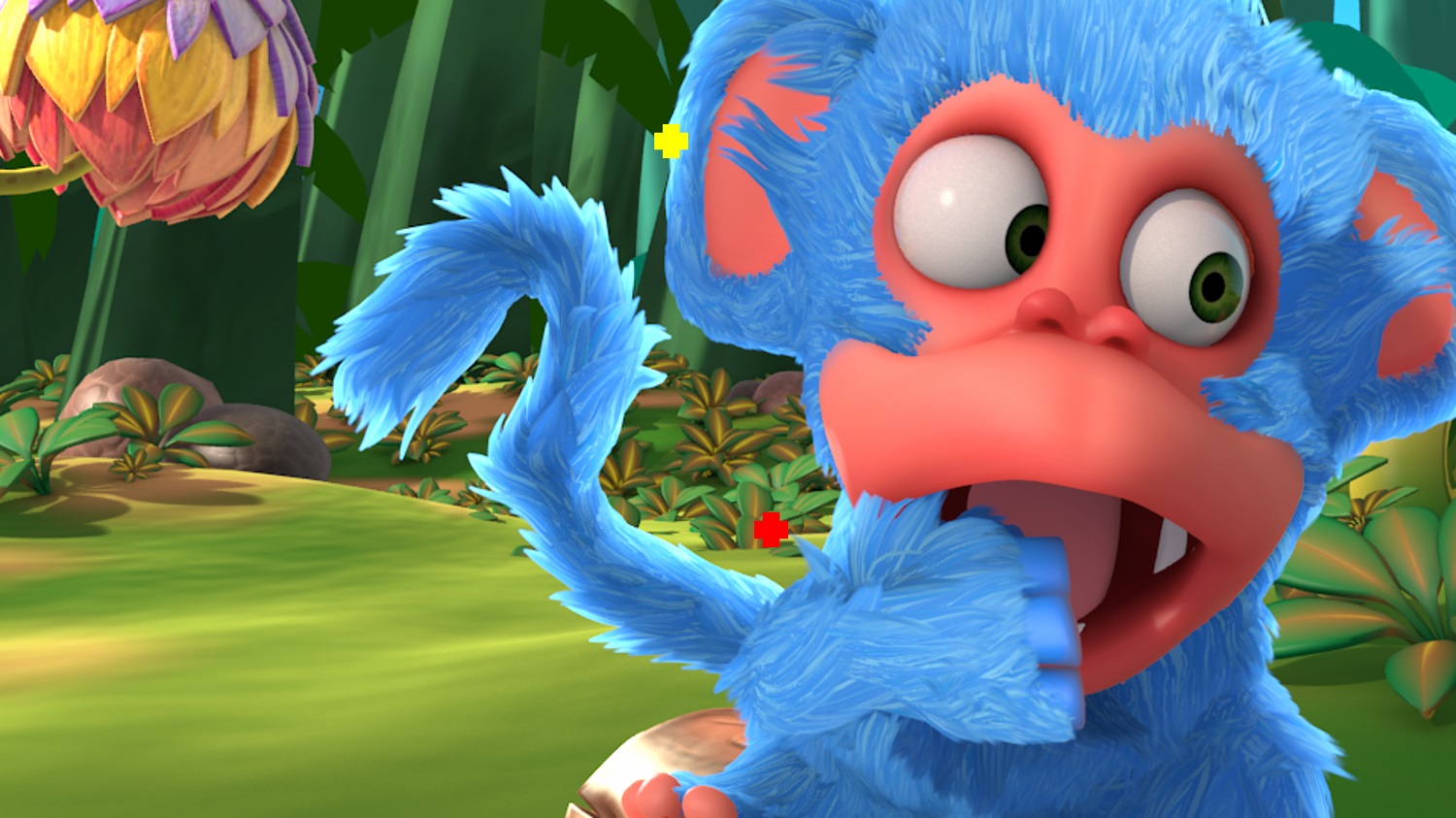}\label{fig1_1}
	}
	\subfloat[Previous Optical Flow Prediction]
	{
		\includegraphics[width=0.48\linewidth]{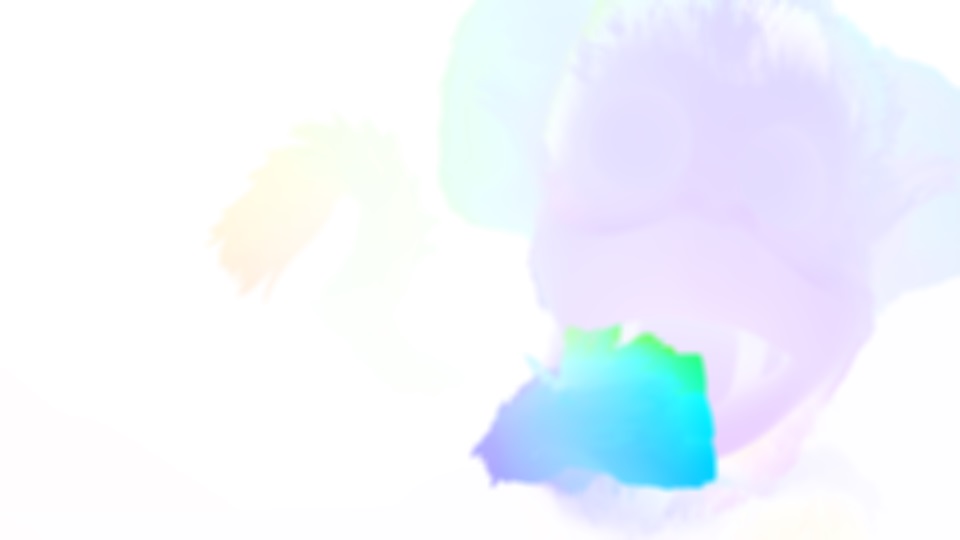}\label{fig1_2}
	}
	\newline
	\subfloat[Warping Based Cost Volume]
	{
		\includegraphics[width=0.48\linewidth]{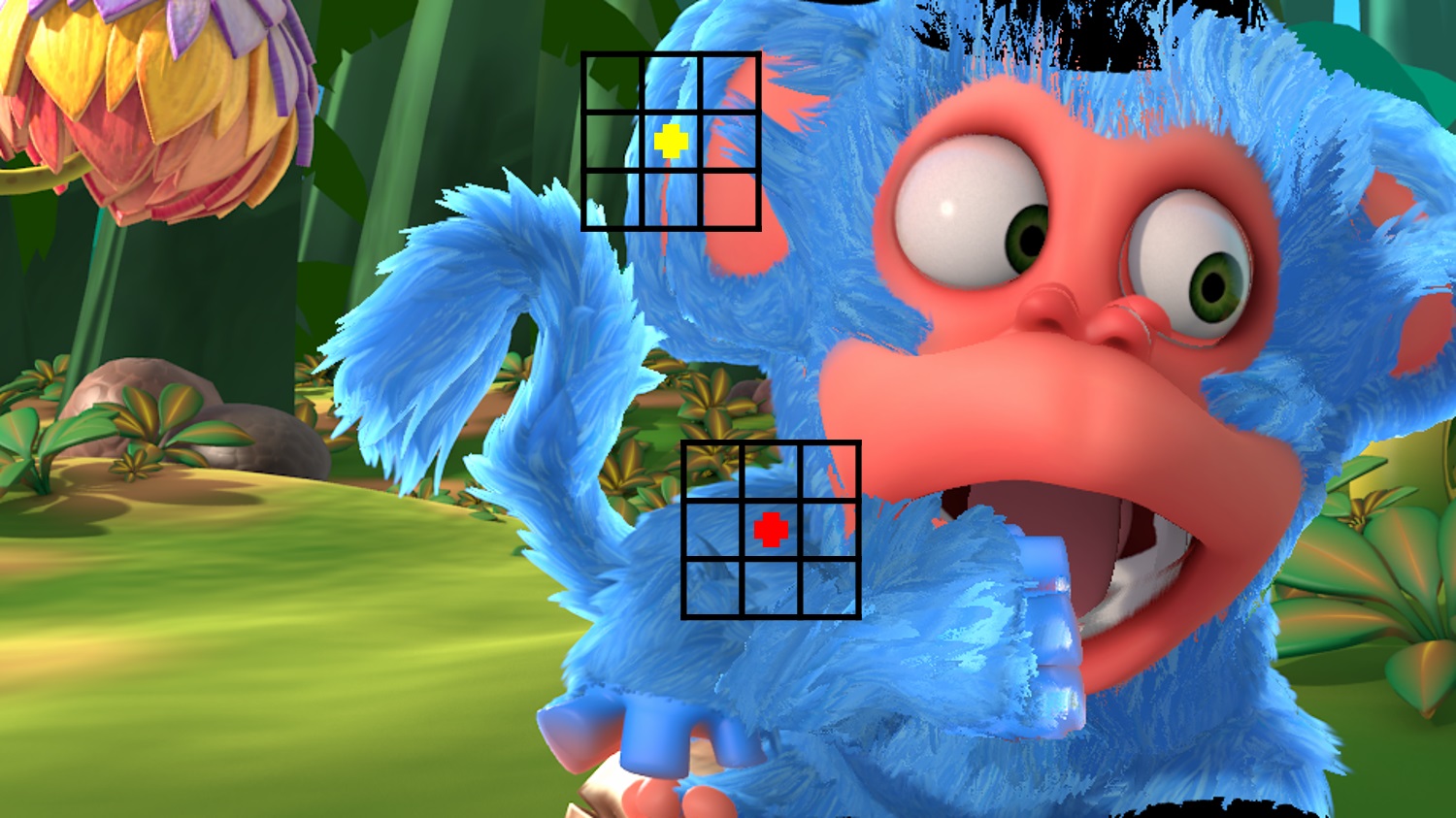}\label{fig1_3}
	}
	\subfloat[Sampling Based Cost Volume]
	{
		\includegraphics[width=0.48\linewidth]{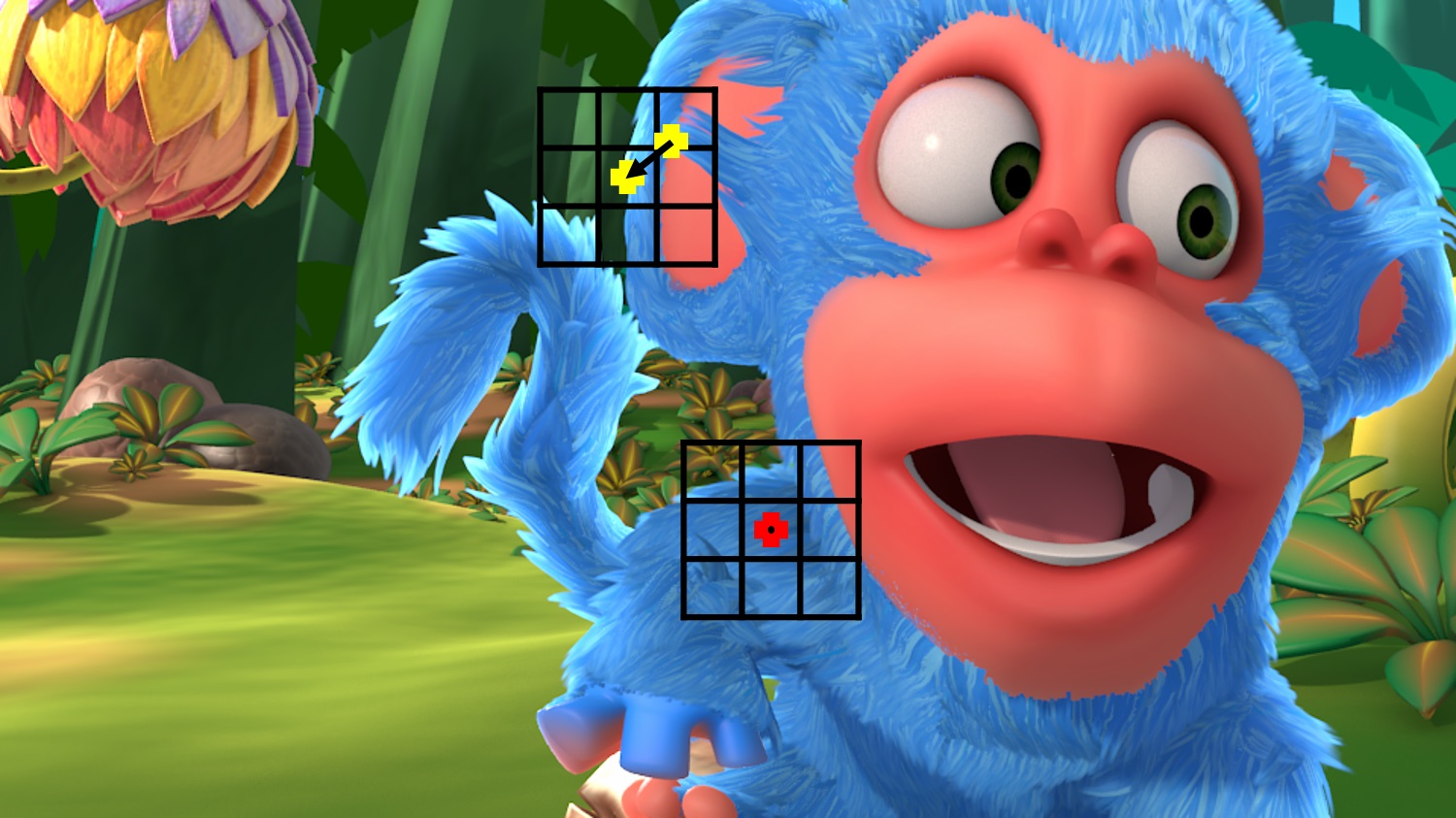}\label{fig1_4}
	}
	\caption{Proposed sampling based cost volume does not have ghosting issue that appears in warping based cost volume.}
	\label{fig1}
	\vspace{-1.0 em}
\end{figure}

With the development of Convolutional Neural Networks (CNNs), deep learning based optical flow networks~\cite{Fischer2015FlowNetLO,Ilg2017FlowNet2E,Sun_2018_CVPR} are proposed together with large synthetic training datasets. Dosovitskiy \textit{et al.}~\cite{Fischer2015FlowNetLO} firstly apply CNNs to optical flow and put forward two networks named FlowNetS and FlowNetC. Though its performance is slightly worse than traditional methods, the running speed is several orders of magnitude faster. The successor FlowNet2~\cite{Ilg2017FlowNet2E} outperforms variational solutions by stacking several basic models and training on multiple datasets with carefully designed learning schedules. However, the large number of parameters hinder it from mobile applications. Concurrently, the liteweight SPyNet~\cite{Ranjan_2017_CVPR} constructs image pyramid and warps the second image to the first one according to prior prediction to estimate residual flow in a coarse-to-fine manner. Nevertheless, it suffers from large performance decline and computation burden.

Recently, PWC-Net~\cite{Sun_2018_CVPR} and LiteFlowNet~\cite{Hui_2018_CVPR} similarly adopt pyramid feature, warping operation and correlation layer in each level to estimate residual flow from coarse to fine. They obtain improvement in both prediction accuracy and model size. Following this, IRR-PWC~\cite{Hur_2019_IRR} proposes shared flow estimators among different scales for iterative residual estimation, which can reduce learnable parameters and speed up convergence. FDFlowNet~\cite{Kong_2020} improves original pyramid structure with a compact U-shape network and proposes efficient partial fully connected flow estimator for fast and accurate inference. Recently, Devon~\cite{Lu_2020_WACV} builds deformable cost volume with multiple dilated rates to capture small fast moving objects and alleviate warping artifacts.

\begin{figure*}[t]
	\centering
	\includegraphics[width=0.9\textwidth]{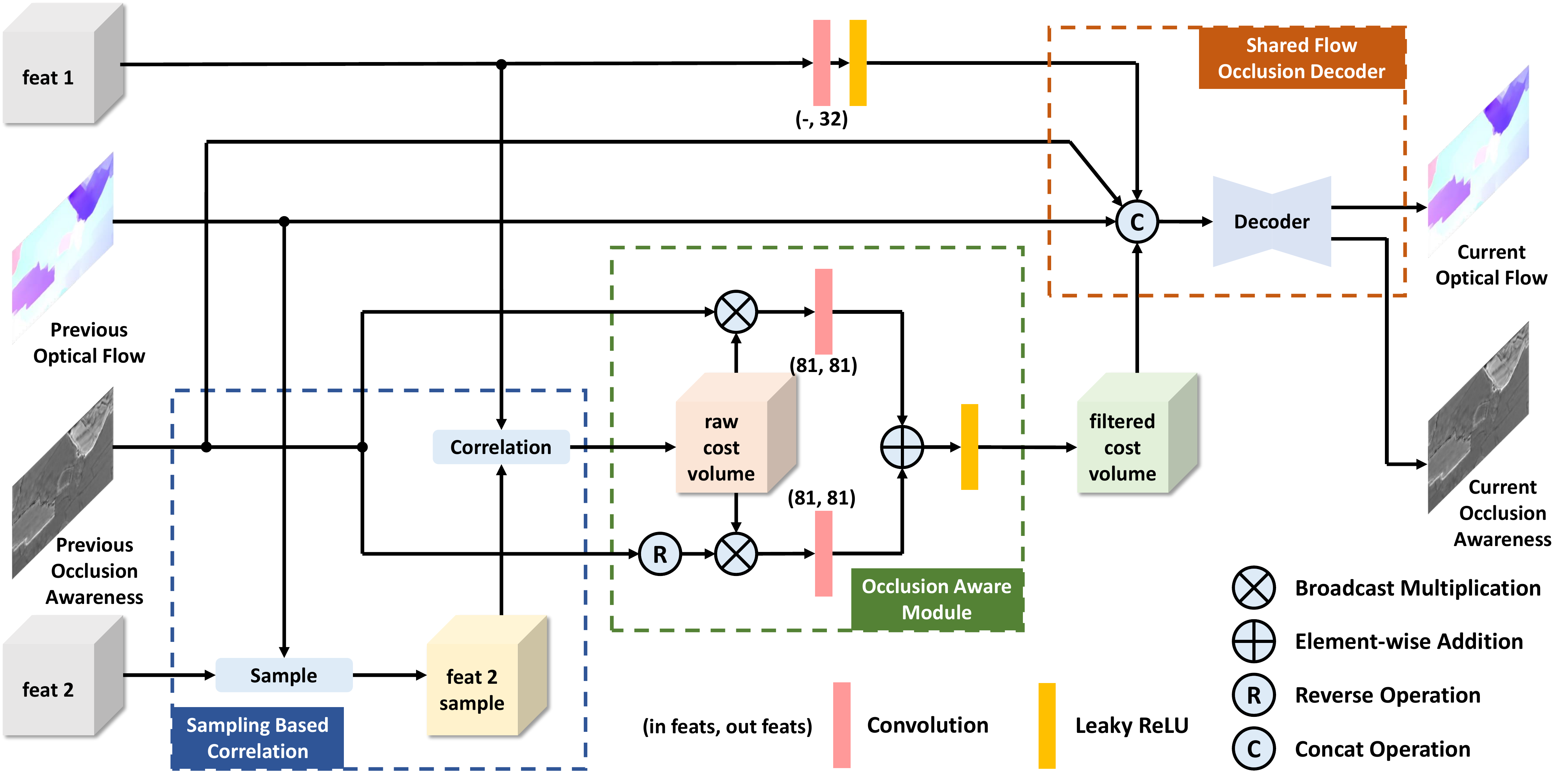}
	\caption{A single pyramid level for occlusion aware optical flow estimation in OAS-Net, feat 1 and feat 2 are pyramid features.}
	\label{fig2}
	\vspace{-0.5 em}
\end{figure*}

Though benchmark results are constantly being promoted, there are still challenges blocking further progresses. Since the high efficiency on reducing large displacement, warping operation appears in almost all advanced flow architectures~\cite{Ilg2017FlowNet2E,Ranjan_2017_CVPR,Sun_2018_CVPR,Hui_2018_CVPR, Hur_2019_IRR,Yin_2019_CVPR,Kong_2020}. However, as shown in Fig~\ref{fig1_3}, the warped target image suffers from ghosting problem, \textit{e.g.}, the yellow cross region, which leads to ambiguity of original scene structure and will further damage performance. Another challenge is the inevitable occlusion, such as the red cross shown in Fig~\ref{fig1}. To solve above problems, we propose a lightweight and efficient occlusion aware sampling network, termed OAS-Net, for accuracy optical flow estimation. We summarize our contributions as follows:

$\bullet$ A novel sampling based cost volume is adopted to avoid ghosting phenomenon and reduce matching ambiguity.

$\bullet$ An occlusion aware module is seamless embedded into the coarse-to-fine flow architecture to endow raw matching cost better occlusion awareness.

$\bullet$ With shared optical flow and occlusion awareness decoder, a new lightweight yet efficient OAS-Net is constructed, which achieves state-of-the-art performance.

\section{Method}
\subsection{Overall Network Architecture}
Given input frames of $I_1, I_2$, our OAS-Net estimates optical flow in a coarse-to-fine manner like many approaches~\cite{Sun_2018_CVPR,Hui_2018_CVPR,Yin_2019_CVPR,Kong_2020}, where a feature extraction network is adopted to build pyramid features $f_1^{k}, f_2^{k}, k=1,2, \ldots, 6$. Each pyramid scale in OAS-Net contains two convolution layers and the first stride 2 convolution down samples spatial resolutions. Commonly, we set an incremental 16, 32, 64, 96, 128 and 160 feature maps from level 1 to level 6 respectively.

As shown in Fig~\ref{fig2}, 5 sub-networks sharing the same structure for occlusion aware optical flow estimation are employed in each level to refine $2\times$ upsampled flow fields together with occlusion awareness maps. For completeness, we first describe the shared optical flow and occlusion decoder. Since there is no dilated context network~\cite{Sun_2018_CVPR, Hur_2019_IRR} utilized by OAS-Net, we set a sequential connected decoder with three more convolution layers to enlarge receptive field for fair comparison, which outputs 128, 128, 128, 128, 128, 96, 64 and 32 channels. The final flow and occlusion maps are predicted by two separate convolution heads making OAS-Net lightweight and efficient. To keep meaningful range of occlusion awareness map, we add a non-linear sigmoid activation to restrict output from 0 to 1. Unlike~\cite{Hur_2019_IRR} which relys on both flow and occlusion ground-truth for co-training, our model only requires optical flow label, while the occlusion awareness is automatically learned by flow supervision.

\subsection{Sampling Based Correlation}
One key step in optical flow approaches is to build cost volume which can provide better correspondence representation than direct convolution feature. To deal with the challenging large displacement, mainstream solution is to leverage bilinear sampling based warping operation~\cite{Ranjan_2017_CVPR,Sun_2018_CVPR,Hui_2018_CVPR,Hur_2019_IRR,Yin_2019_CVPR} to reduce offset between source and target corresponding feature according to the roughly estimated flow field. When the target feature is warped to the source one, followed matching cost volume can be built by calculating feature similarity within a local searching area. As shown in Fig~\ref{fig1_3}, this warping based cost volume can be formulated as:
\begin{equation}
c^k({\bf x}, {\bf d}) = {f_1^k({\bf x})} \cdot {f}_{warp}^k({\bf x}+{\bf d}).
\label{eq1}
\end{equation}
Where $f_{warp}^{k}$ means warped target feature in level $k$, $\bf x$ means spatial coordinates and $\bf d$ represents searching offsets. $\cdot$ is to calculate innner product for similarity measurement.

As mentioned above, the warping based cost volume suffers from duplicate artifacts, which is also known as ghosting. To solve the problem, we build cost volume with a novel sampling based approach as shown in Fig~\ref{fig1_4} that avoids warping and ghosting issues. This operation can be formulated as:
\begin{equation}
c^k({\bf x}, {\bf d}) = {f_1^k({\bf x})} \cdot {f}_2^k({\bf x}+{\bf f}+{\bf d}).
\label{eq2}
\end{equation}
Where the new vector $\bf f$ denotes previous optical flow field in the location of coordinate $\bf x$. Different from a dilated differential volume in~\cite{Lu_2020_WACV}, we adopt inner product based volume whose $\bf d$ traverses a searching square with radius 4. Compared to Eq~\ref{eq1}, the flow based warping operation is replaced by directly sampling flow guided searching grids in the target pyramid feature. Experiments will show proposed sampling based correlation method helps to improve performance.

\subsection{Occlusion Aware Module}
Although sampling based cost volume can bypass annoying ghosting feature, another challenging occlusion problem is unavoidable. As shown in Fig~\ref{fig1}, the red cross pointing bush in the source image is covered by the moving arm in the target image. This type of occluded region is unrecoverable no matter to take either correlation methods.

In regard to supervised methods~\cite{Fischer2015FlowNetLO,Ilg2017FlowNet2E,Ranjan_2017_CVPR,Sun_2018_CVPR,Hui_2018_CVPR,Yin_2019_CVPR,Kong_2020,Lu_2020_WACV}, little attention has been paid to handle occlusion explicitly during pure supervison by densely labeled optical flow. To endow flow estimation with occlusion awareness, we present a novel occlusion aware module to better handle two types of matching regions in raw cost volume separately, which can be seamlessly embedded into the coarse-to-fine structure.

As depicted in Fig~\ref{fig2}, there are four steps to build occlusion aware cost volume. First, previous $2\times$ upsampled occlusion awareness is subtracted from a full one tensor to create a reversed occlusion awareness map. Second, the complementary occlusion maps separately multiply raw cost volume to extract two occlusion reweighted cost volumes. Third, two disparate convolution layers are employed to filter above occlusion aware cost volumes respectively. Finally, two filtered matching costs are merged by addition following a leaky ReLU layer. The whole process can be formulated as:
\begin{equation}
\begin{split}
c_{oa}^k({\bf x}, {\bf d}) = \; & {\rm lrelu}({\rm conv_{1}}(O({\bf x}) \otimes c^k({\bf x}, {\bf d})) \\
& \oplus {\rm conv_{2}}(({\bf 1} - O({\bf x})) \otimes c^k({\bf x}, {\bf d}))).
\label{eq3}
\end{split}
\end{equation}
Where $O(\bf x)$ means occlusion awareness map, $\otimes$ represents broadcast multiplication and $\oplus$ stands for element-wise addition. Referring to Eq~\ref{eq3}, proposed occlusion awareness map can also be interpreted as occlusion probability map, which can be viewed as one type of self-attention mechanism~\cite{NIPS2017_7181}. We will visualize occlusion awareness and demonstrate our occlusion aware cost volume can improve optical flow accuracy, especially in fast-moving scenarios.

\section{Experiments}
\subsection{Training Details}
We adopt the two stage training schedule in FlowNet2~\cite{Ilg2017FlowNet2E}. OAS-Net is firstly trained on FlyingChairs dataset~\cite{Fischer2015FlowNetLO} with $S_{short}$ learning schedule, \textit{i.e.}, initial learning rate is set to $1e-4$ and decays half at $300k, 400k$ and $500k$ iterations with total $600k$ iterations. Then, the pretrained model is fine-tuned on FlyingThings3D dataset~\cite{MIFDB16} following the $S_{fine}$ schedule, \textit{i.e.}, with initial learning rate $1e-5$ and decays half at $200k, 300k$ and $400k$ iterations with total $500k$ iterations. For FlyingChairs, batch size is set to 8 and crop size is set to $320 \times 448$, and for FlyingThings3D, we take batch size 4 and adopt $384 \times 768$ crops. Multiple data augmentation methods, including mirror, translate, rotate, zoom, squeeze and color jitter are employed to enrich training distribution and prevent overfitting. Since pyramid structure of OAS-Net is the same with PWC-Net~\cite{Sun_2018_CVPR} and LiteFlowNet~\cite{Hui_2018_CVPR}, we use the same multi-scale L2 loss~\cite{Sun_2018_CVPR} for supervised learning. Adam~\cite{Kingma_Adam} optimizer is adopted in all stages, and our experiments are conducted on one NVIDIA GTX 1080Ti GPU with PyTorch.

\subsection{Ablation Study}
\begin{table}[!t]
	\renewcommand\arraystretch{1.0}
	\normalsize
	\caption{Ablation study of proposed modules.}
	\vspace{-1.5 em}
	\begin{center}
		\centering
		\setlength{\tabcolsep}{3mm}{
			\begin{tabular}{c|c|c|c}
				\toprule
				Correlation & Occlusion & Sintel & KITTI \\
				Method & Awareness & Final & 2012 \\
				\midrule
				Warping & & 4.05 & 4.62 \\
				Warping & \Checkmark & 3.98 & 4.37 \\
				Sampling & & 3.86 & 4.44 \\
				Sampling & \Checkmark & {\bf 3.79} & {\bf 4.11} \\
				\bottomrule
		\end{tabular}}
	\end{center}
	\label{tab1}
	\vspace{-1.0 em}
\end{table}

To explore and verify value of proposed approaches, we conduct ablation with 4 variations based on pairwise combinations. For sake of effective comparison, all variants follow above two stage training schedule and are evaluated on Sintel Final, KITTI 2012 training datasets, as listed in Table~\ref{tab1}.

\begin{table*}[t]
	\renewcommand\arraystretch{1.0}
	\normalsize
	\caption{Comparison on Sintel and KITTI benchmarks. Optical flow accuracy is measured by end point error.}
	\vspace{-1.5 em}
	\begin{center}
		\centering
		\setlength{\tabcolsep}{5mm}{
			\begin{tabular}{r|cc|cc|cc|c|c}
				\toprule
				\multicolumn{1}{c|}{\multirow{2}{*}{Method}} & \multicolumn{2}{|c|}{Sintel Clean} & \multicolumn{2}{|c|}{Sintel Final} & \multicolumn{2}{|c|}{KITTI 2012} & \multicolumn{1}{|c}{Parameters} & \multicolumn{1}{|c}{Time} \\
				& train & test & train & test & train & test & (M) & (s) \\
				\midrule
				FlowNetC~\cite{Fischer2015FlowNetLO} & 4.31 & 6.85 & 5.87 & 8.51 & 9.35 & - & 39.18 & 0.050 \\
				SPyNet~\cite{Ranjan_2017_CVPR} & 4.12 & 6.64 & 5.57 & 8.36 & 9.12 & 4.7 & {\bf 1.20} & 0.055 \\
				FlowNet2~\cite{Ilg2017FlowNet2E} & {\bf 2.02} & 4.16 & {\bf 3.14} & 5.74 & 4.09 & 1.8 & 162.49 & 0.120 \\
				LiteFlowNet~\cite{Hui_2018_CVPR} & 2.48 & 4.54 & 4.04 & 5.38 & {\bf 4.00} & 1.6 & 5.37 & 0.055 \\
				PWC-Net~\cite{Sun_2018_CVPR} & 2.55 & 4.39 & 3.93 & 5.04 & 4.14 & 1.7 & 8.75 & 0.035 \\
				IRR-PWC~\cite{Hur_2019_IRR} & - & 3.84 & - & {\bf 4.58} & - & 1.6 & 6.36 & 0.150 \\
				HD3-Flow~\cite{Yin_2019_CVPR} & 3.84 & 4.79 & 8.77 & 4.67 & 4.65 & {\bf 1.4} & 39.56 & 0.080 \\
				Devon~\cite{Lu_2020_WACV} & - & 4.34 & - & 6.35 & - & 2.6 & - & 0.050 \\
				OAS-Net (Ours) & 2.55 & {\bf 3.65} & 3.79 & 5.01 & 4.11 & {\bf 1.4} & 6.16 & {\bf 0.030} \\
				\bottomrule
		\end{tabular}}
	\end{center}
	\label{tab2}
	\vspace{-1.0 em}
\end{table*}

Compared with the baseline method, which adopts warping based correlation and does not include occlusion awareness module, only replacing warping based cost volume with sampling based one reduces end point error of $4.7\%$ on Sintel Final and $3.9\%$ on KITTI 2012. While only adding occlusion awareness module reduces error of $1.7\%$ on Sintel and $5.4\%$ on KITTI. It concludes that sampling is more remarkable on Sintel which contains more non-rigid motion and has higher requirement on distinct scene structure, while occlusion awareness is more effective on KITTI that involves large movement and occluded regions. Finally, combining these two approaches gets best results that reduces end point error of $6.4\%$ on Sintel Final and $11.0\%$ on KITTI 2012, demonstrating our contributions are cooperative and complementary.

\subsection{Benchmark Results}
\begin{figure}[!t]
	\captionsetup[subfigure]{farskip=1pt}
	\centering
	\subfloat[First Image]
	{
		\includegraphics[width=0.48\linewidth]{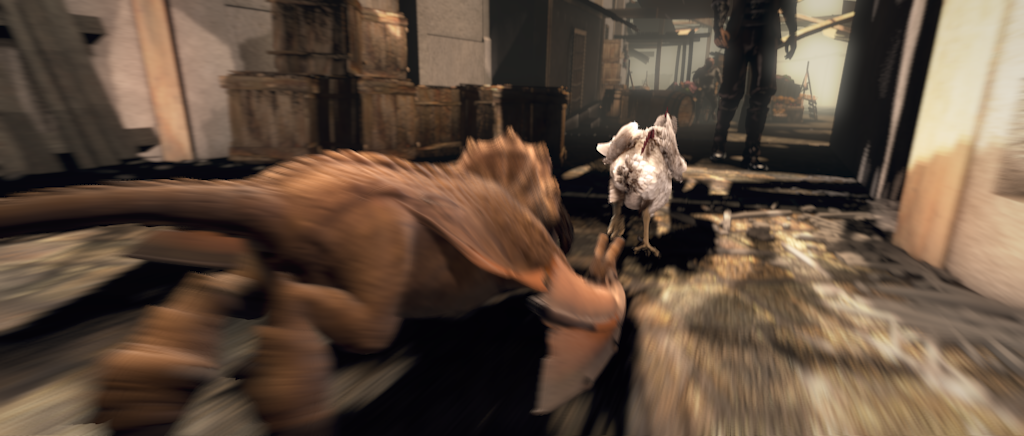}\label{fig3_1}
	}
	\subfloat[Ground Truth]
	{
		\includegraphics[width=0.48\linewidth]{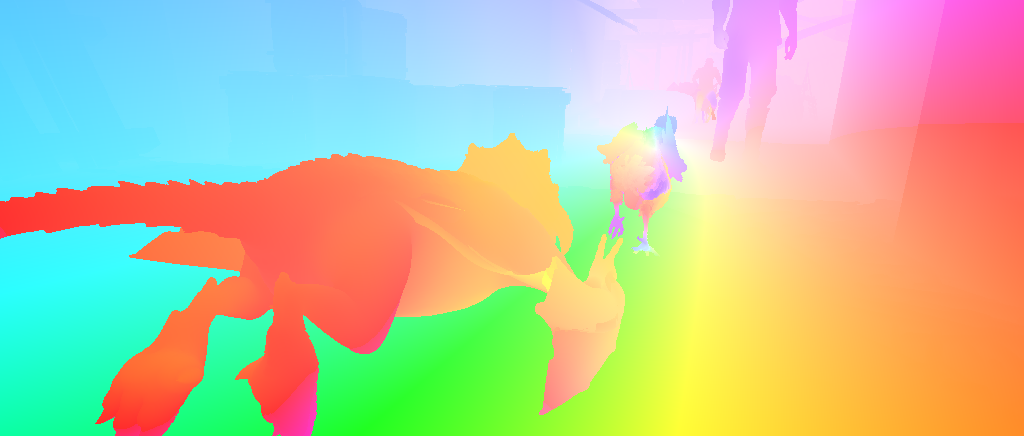}\label{fig3_2}
	}
	\newline
	\subfloat[PWC-Net~\cite{Sun_2018_CVPR}]
	{
		\includegraphics[width=0.48\linewidth]{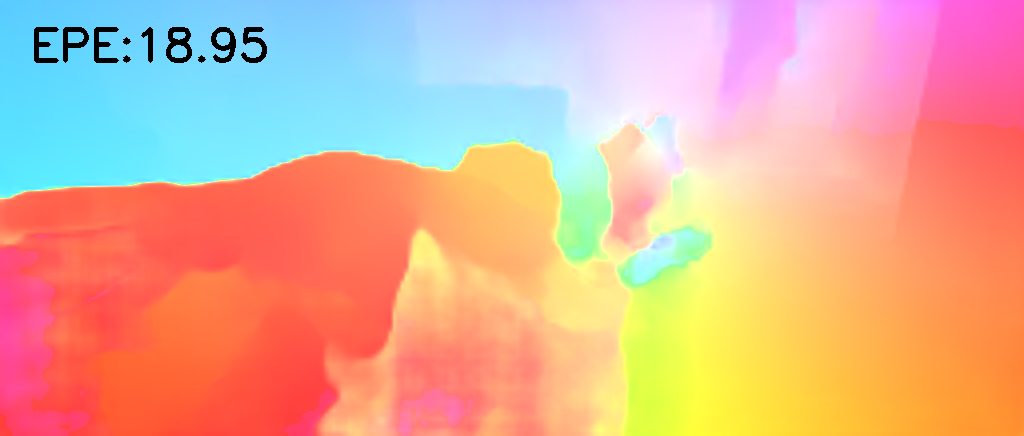}\label{fig3_3}
	}
	\subfloat[FlowNet2~\cite{Ilg2017FlowNet2E}]
	{
		\includegraphics[width=0.48\linewidth]{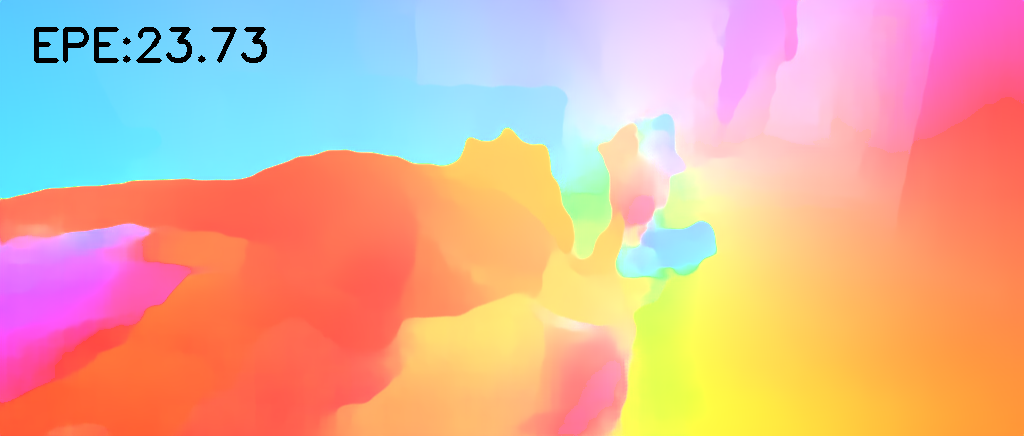}\label{fig3_4}
	}
	\newline
	\subfloat[OAS-Net]
	{
		\includegraphics[width=0.48\linewidth]{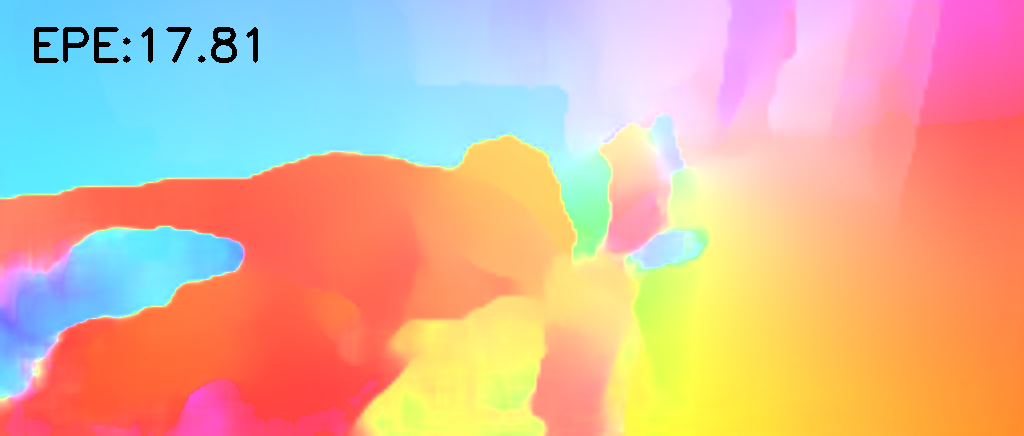}\label{fig3_5}
	}
	\subfloat[Occlusion Awareness]
	{
		\includegraphics[width=0.48\linewidth]{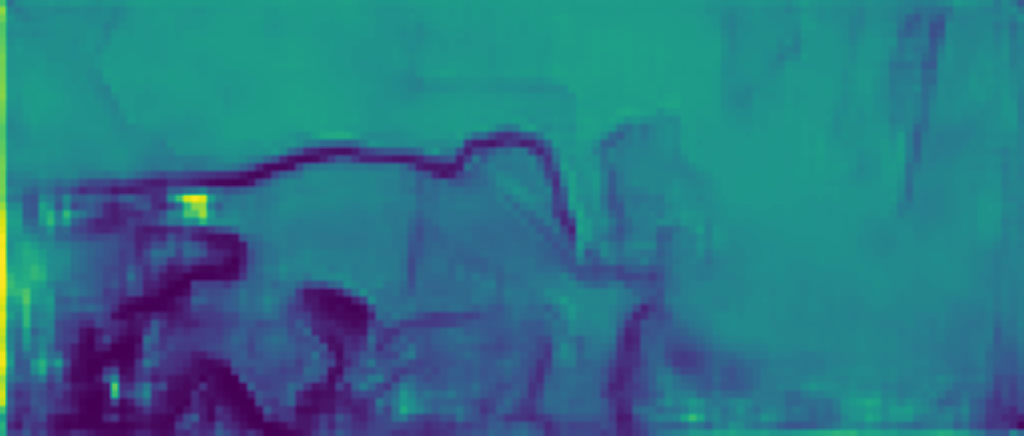}\label{fig3_6}
	}
	\caption{Comparison on Sintel Final test dataset.}
	\label{fig3}
	\vspace{-1.0 em}
\end{figure}

To compare with state-of-the-art methods~\cite{Ilg2017FlowNet2E,Hui_2018_CVPR,Sun_2018_CVPR,Hur_2019_IRR,Yin_2019_CVPR}, we first evaluate above two stage trained OAS-Net on Sintel~\cite{Butler:ECCV:2012} and KITTI~\cite{Geiger2012CVPR} training sets. For comparison on Sintel test benchmark, we fine-tune OAS-Net on mixed Sintel datasets, where we adopt batch size 4, with 2 from Clean, and 2 from Final. Similar to IRR-PWC~\cite{Hur_2019_IRR}, learning rate is initially set to $5e-5$ and disturbed every $150k$ iterations over total $600k$ iterations. To evaluate on KITTI 2012 test benchmark, we fine-tune OAS-Net on mixed KITTI 2012 and KITTI 2015 training sets following the same schedule on Sintel while reducing amplitude of spatial augmentation. All results together with model size and running speed (measured on Sintel resolution) are listed in Table~\ref{tab2}.

\begin{figure}[!t]
	\captionsetup[subfigure]{farskip=1pt}
	\centering
	\subfloat[First Image]
	{
		\includegraphics[width=0.48\linewidth]{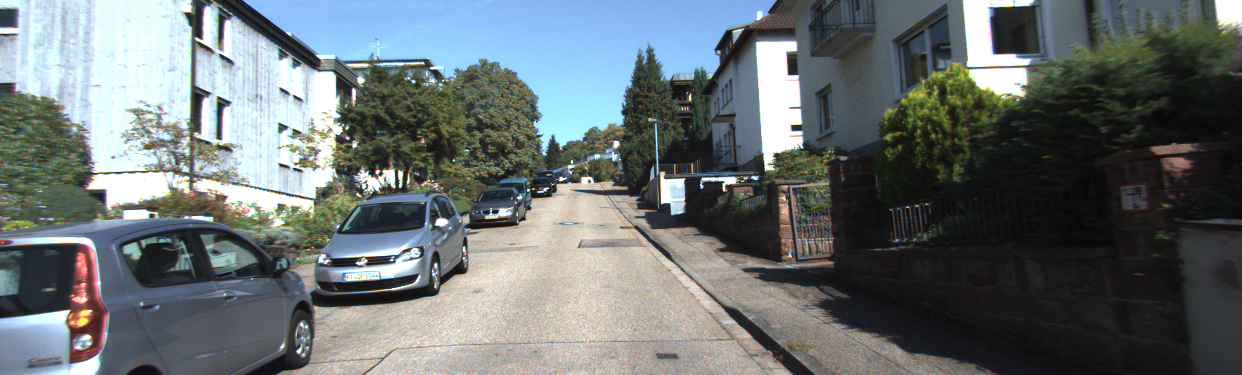}\label{fig4_1}
	}
	\subfloat[Ground Truth]
	{
		\includegraphics[width=0.48\linewidth]{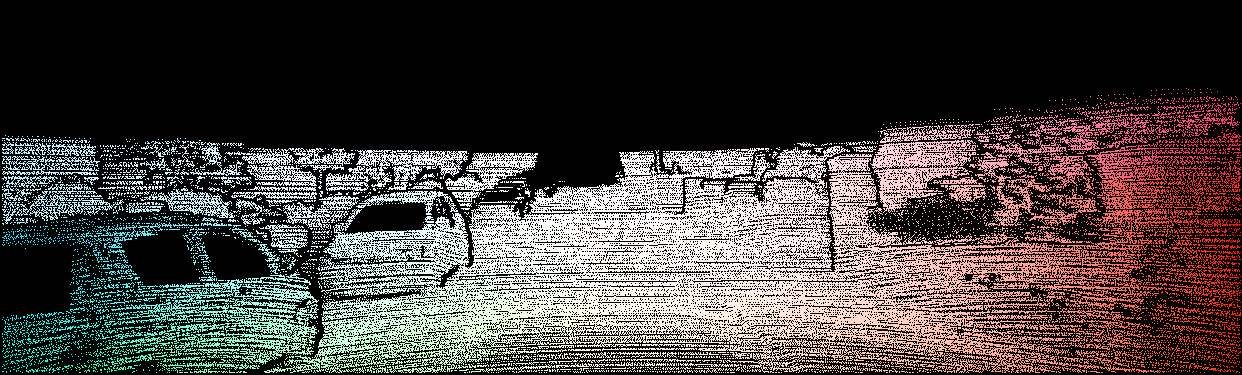}\label{fig4_2}
	}
	\newline
	\subfloat[Optical Flow]
	{
		\includegraphics[width=0.48\linewidth]{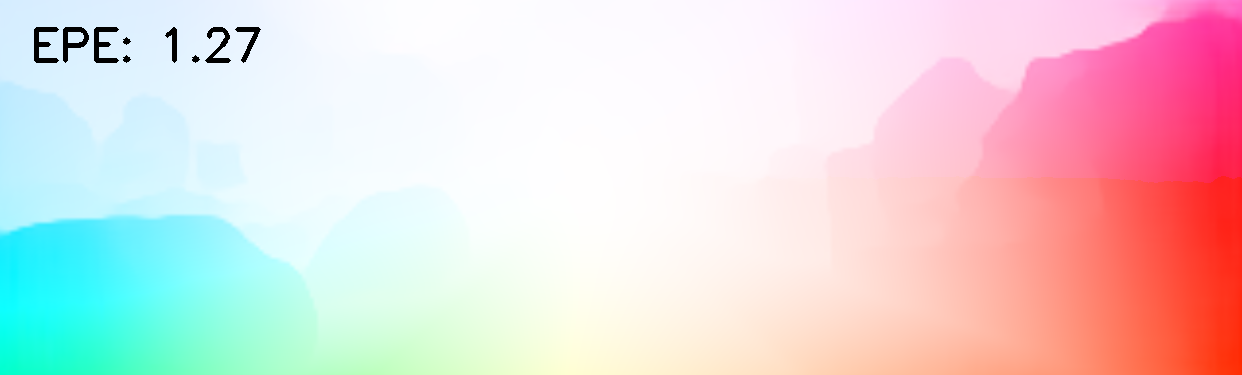}\label{fig4_3}
	}
	\subfloat[Occlusion Awareness]
	{
		\includegraphics[width=0.48\linewidth]{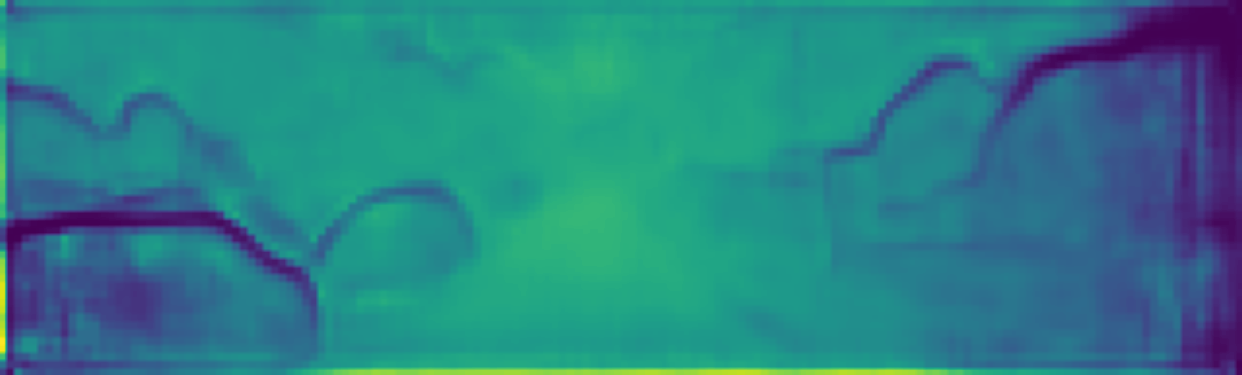}\label{fig4_4}
	}
	\caption{Visual example of OAS-Net on KITTI dataset.}
	\label{fig4}
	\vspace{-0.5 em}
\end{figure}

OAS-Net performs better than all the other approaches on Sintel Clean test dataset, it improves about $5\%$ than the second one. OAS-Net also surpasses PWC-Net~\cite{Sun_2018_CVPR} on Sinel Final test set slightly, but is exceeded by IRR-PWC~\cite{Hur_2019_IRR} and HD3-Flow~\cite{Yin_2019_CVPR}. However, our model has less learnable parameters, \textit{i.e.}, $6.16$M vs $6.36$M$/$$39.56$M and runs several times faster, \textit{i.e.}, $0.03$s vs $0.15$s$/$$0.08$s. We further evaluate it on a more challenging real-world KITTI dataset. As shown in Table~\ref{tab2}, OAS-Net achieves the same best result as HD3-Flow~\cite{Yin_2019_CVPR} on KITTI 2012 test benchmark with end point error of $1.4$, which reduces $12.5\%$ error than LiteFlowNet~\cite{Hui_2018_CVPR}.

To verify our approaches visually, we show one example on each Sintel Final test and KITTI training sets, as depicted in Fig~\ref{fig3} and Fig~\ref{fig4}. It can be seen that our occlusion awareness maps have correctly emphasized the probable occluded regions, and help to improve optical flow accuracy, such as the left bottom corner of Fig~\ref{fig3} and fast moving edges in Fig~\ref{fig4}.

\section{Conclusion}
In this paper, we have presented a new OAS-Net for accuracy optical flow estimation. To keep clear scene structure, we propose to use sampling based correlation instead of noisy warping method. Then, we embed novel occlusion aware module into the coarse-to-fine flow architecture to endow raw cost volume better occlusion awareness. Finally, our optical flow and occlusion awareness share the same decoder making OAS-Net lightweight and fast. Experiments on both synthetic Sintel and real-world KITTI datasets demonstrate the effectiveness of proposed approaches and show its state-of-the-art results on challenging benchmarks.

\bibliographystyle{IEEEbib}
\bibliography{refs}

\end{document}